\documentclass[runningheads,orivec]{llncs}
\usepackage[T1]{fontenc}
\usepackage{graphicx,verbatim}
\usepackage{multirow}
\usepackage{amsmath}       
\usepackage{amssymb}       
\usepackage{amsfonts}      
\usepackage{mathtools}     
\usepackage{bm}            
\usepackage{enumitem}      
\usepackage{hyperref}      
\usepackage{cite}          
\usepackage{booktabs}

\begin{document}
\title{Multivariate Gaussian NeRF for Wide Field-of-View Ultrasound  Reconstruction}
%
\author{Patris Valera\inst{1,2} \thanks{These authors contributed equally to this work.} \and
Magdalena Wysocki\inst{1} \textsuperscript{*} \thanks{Corresponding author}\and 
Felix Duelmer \inst{1, 2} \and 
Mohammad Farid Azampour \inst{1, 2}  \and
Sebastian Herz \inst{3} \and
Stefan Wörz \inst{3} \and
Nassir Navab \inst{1,2} }
%
%
\institute{
Chair for Computer Aided Medical Procedures (CAMP), Technical University of Munich, Germany \\
\email{\{patris.valera, magdalena.wysocki, felix.duelmer, mf.azampour, nassir.navab\}@tum.de}
\and
Munich Center for Machine Learning (MCML), Munich, Germany \and
LUMA Vision, Beech Hill Road, Dublin, Ireland \\
\url{http://www.lumavision.com} \\
\email{\{sebastian.herz, stefan.woerz\}@lumavision.com} \\
}

\maketitle              
\sloppy
\begin{abstract}
Wide Field-of-View (WFoV) reconstruction enhances 3D ultrasound imaging by providing valuable anatomical context for segmentation models and visualization. Clinical ultrasound volumes are predominantly acquired using convex probes, which generate expanding, diverging acoustic beams to maximize anatomical coverage. Stitching these sweeps together traditionally introduces significant compounding artifacts and aliasing due to depth-dependent resolution changes.
Here, we introduce Ultra-Wide-NeRF, a Multivariate 3D Gaussian (MVG) NeRF-based method for WFoV ultrasound reconstruction. By explicitly modeling the complex beam geometry using distance-dependent convex volumetric sampling and anisotropic 3D Gaussians, our method inherently mitigates these compounding artifacts and provides anti-aliasing. 
Beyond simply reconstructing a static 3D grid, our NeRF-based approach yields a continuous neural representation of the tissue, enabling the synthesis of high-fidelity novel views from arbitrary virtual trajectories. 
We validate Ultra-Wide-NeRF for intracardiac echocardiography on phantom and porcine datasets, demonstrating that our method expands the spatial context important in intraoperative navigation.
Code will be open-sourced upon publication.

\end{abstract} 
\section{Introduction}
Visualizing large, complex anatomical structures in their entirety is critical for accurate clinical diagnosis and intraoperative navigation. 
While three-dimensional (3D) ultrasound (US) provides a real-time, accessible, and radiation-free imaging modality~\cite{fenster2001three, hsu2012real}, individual acquisitions are physically constrained to a limited Field-of-View (FoV). To overcome this restriction and achieve the necessary macroscopic anatomical context, multiple overlapping 3D US acquisitions can be stitched together using spatial compounding. This process yields a Wide FoV (WFoV) panoramic volume that captures extended anatomical regions.
Conventional WFoV compounding methods align overlapping 3D sequences using feature- or intensity-based registration~\cite{poon2006three,flach2016pure,hickler2024panoramic}, which inevitably introduces interpolation artifacts. Recent Neural Radiance Fields (NeRFs) for ultrasound provide a continuous, physics-aware alternative. Approaches like Ultra-NeRF~\cite{wysocki2024ultra}, NeRF-US~\cite{dagli2024nerfus}, and UlRe-NeRF~\cite{guo2024ulre} leverage implicit representations mapped via ray tracing~\cite{salehi2015patient,burger2012real,duelmer2025ultraray}. By evaluating the entire acoustic propagation path rather than assigning static voxel intensities, these models inherently avoid traditional interpolation artifacts. Despite these advantages, however, current ultrasound NeRF models rely on equidistant point sampling along infinitesimally thin rays. This formulation fundamentally fails for the diverging geometries of clinical convex probes, where the physical volume sampled by the acoustic pulse expands proportionally with depth~\cite{cobbold2006foundations, prince2006medical}. Ignoring this expanding sample volume causes far-field aliasing, depth-dependent resolution loss, and unaddressed anisotropy across the spatial axes. 

Here, we introduce Multivariate Gaussian NeRF for Wide Field-of-View Ultrasound Reconstruction (Ultra-Wide-NeRF) to unlock the full potential of neural rendering in ultrasound. Bridging NeRFs with convex geometry, our framework aligns the sampling strategy directly with how the diverging ultrasound beam behaves in space. By replacing standard thin-ray approximations with distance-dependent convex volumetric sampling and multivariate Gaussians, we explicitly model the expanding spatial footprint of the acoustic beam. This continuous formulation naturally captures depth-dependent resolution, acting as an inherent anti-aliasing filter that eliminates traditional stitching artifacts. Consequently, Ultra-Wide-NeRF successfully fuses multiple sweeps into a seamless panoramic volume while enabling realistic novel view synthesis from unobserved trajectories. 
Validated on intracardiac echocardiography (ICE), our framework delivers comprehensive spatial context in two distinct ways: NeRF-based panoramic reconstruction effectively reduces stitching artifacts and enhances anatomical borders to extend context along the acquired trajectory, while high-fidelity novel view synthesis generates views from unobserved virtual perspectives. 
Ultimately, this combined, high-quality panoramic reconstruction establishes a robust foundation of visual context for seamless intraoperative cardiac navigation.
\section{Methodology} 

Our objective is to reconstruct a unified, WFoV representation from a sequence of $N$ overlapping, tracked 3D US volumes $\{\mathcal{V}_i \in \mathbb{R}^{D \times H \times W}\}_{i=1}^N$ with corresponding poses $\{\mathbf{P}_i \in SE(3)\}_{i=1}^N$, where $D,H,W$ are the volume dimensions.
As shown in Fig. \ref{fig:overview}, the proposed method optimizes a continuous physics-informed neural field and uses Multivariate 3D Gaussians (Sec. \ref{ssec:3dgauss}) to define a WFoV US volume $\mathcal{PAN} \in \mathbb{R}^{(D \times N)  \times H \times W}$ that spans the entire catheter trajectory.
We define the coordinate system such that the $z$-axis is aligned with the catheter length (elevational direction, or volume-depth $D$), the $y$-axis with the lateral direction of the ultrasound beam, and the $x$-axis with the axial (scan-depth) direction. We refer to $z$ as the depth direction in the volumetric sense.




\begin{figure}[ht] 
    \centering
    \includegraphics[width=\textwidth]{gfx/overview_new.png} 
    \caption{\textbf{Overview of the proposed method for unified, WFoV Volume Reconstruction}. We model a convex transducer by radially shooting rays from an opening angle $\Theta$ which are then mapped to 3D Cartesian world space via their associated poses. Afterwards, \textbf{Ultra-Wide-NeRF} estimates a physics-informed continuous radiance field by fitting multivariate 3D Gaussians along the three resolution axes with an explicit attenuation model. Ultra-Wide-NeRF generates both WFoV Volume Reconstruction and novel view synthesis from unseen probe poses.}
    \label{fig:overview}
\end{figure}

\subsection{Physics-Based Neural Ultrasound Fields}

In NeRF's for ultrasound  a coordinate-based network maps a 3D location $\mathbf{x} \in \mathbb{R}^{3}$ to a vector of acoustic parameters $\tau \in \mathbb{R}^{d}$, which serves as an input to an acoustic forward synthesis model. This model computes voxel intensity at the location $\mathbf{x}$ assuming a linear transducer casting parallel rays $\mathbf{r}(t) = \mathbf{o} + t\mathbf{d}$ sampled at discrete points $t$ (Fig.~\ref{fig:sampling_strategies}). This sampling strategy is not optimal for convex and volumetric probes since the physical gaps between scanlines expand with depth, causing severe far-field aliasing. 

\subsection{Convex Probe Geometry}\label{ssec:convexprobe}
\begin{figure}[ht]
    \centering
    \includegraphics[width=\linewidth]{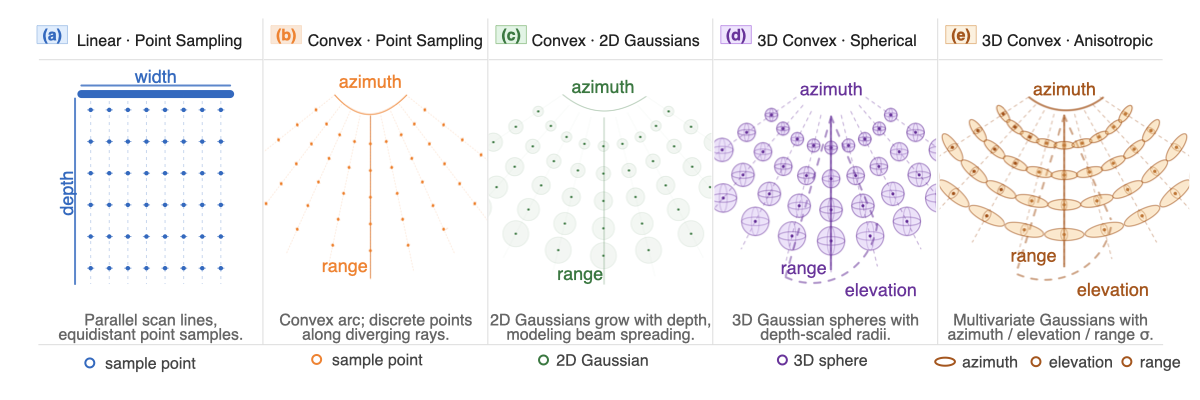}
\caption{\textbf{Ultrasound beam sampling strategies.} \textbf{(a)} Parallel point sampling for linear arrays. \textbf{(b)} Diverging point sampling for convex arrays, highlighting depth-dependent under-sampling. \textbf{(c)} 2D Gaussian sampling, capturing the in-plane profile. \textbf{(d)} Isotropic 3D Gaussian sampling, adding the elevation footprint. \textbf{(e)} Proposed multivariate 3D Gaussian sampling, accurately modeling the anisotropic volumetric expansion of diverging beams.}
    \label{fig:sampling_strategies}
\end{figure}
For the \textbf{ 2D convex FoV }, the domain is defined in polar coordinates $(r, \theta)$ where $r$ defines range and $\theta$ defines azimuth on an image plane $\Phi_i$ corresponding to the pose $\mathbf{P}_i$.
The convex geometry is parameterized by an inner radius $R_{in}$, an outer radius $R_{out}$, and an opening angle $\Theta \in (0^\circ, 360^\circ]$. 
In a simple case (Fig~\ref{fig:sampling_strategies}) the rays are cast uniformly over $\theta \in [-\frac{\Theta}{2}, \frac{\Theta}{2}]$ and sampled along
$r \in [R_\mathrm{in}, R_\mathrm{out}]$. 
To evaluate these samples within the NeRF architecture, polar samples are transformed into Cartesian coordinates $(x, y, z) := (r \sin(\theta), r \cos(\theta) - R_{in}, z) $ and positioned in $\mathbb{R}^{3}$ using the pose $\mathbf{P}_i$.

To extend the convex probe geometry to 3D, we adopt a multi-planar acquisition model rather than an explicit spherical parameterization.
Each volume $\mathcal{V}_i$ is formed by stacking $D$ fan-beam (convex) slices, each defined on its own image plane $\Phi_{i,d}$ with associated pose $\mathbf{P}_{i,d}$, where $d \in [0,D)$.
Within each slice, rays are cast in polar coordinates $(r, \theta)$ as described above and lifted into 3D Cartesian world coordinates via a two-step mapping: first, the in-plane polar-to-Cartesian conversion and second, the rigid-body transform $\Phi_{i,d}$ that positions each slice plane in the shared world frame.
The elevational dimension is implicitly defined by the inter-slice pose differences $\|\mathbf{P}_{i,d+1} - \mathbf{P}_{i,d}\|$, which encode the angular or translational sweep of the acquisition.
All ray samples ultimately query the neural field in Cartesian world coordinates, enabling seamless integration of multi-planar acquisitions within a single continuous volumetric representation.


\subsection{From Points to Multivariate 3D Gaussians}\label{ssec:3dgauss}
Approximating the interaction between an acoustic wave and a tissue scatterer as a discrete point estimate is fundamentally ill-posed. This approach disregards the finite volume each sample actually represents, inevitably leading to aliasing.
Hence, we model each sampled region as a conical frustum parameterized by a 3D multivariate Gaussian to create a continuous space. Since 3D US imaging exhibits fundamental anisotropy in both lateral (in-plane) and elevational (cross-plane) directions, we extend the mip-NeRF's~\cite{barron2021mip} frustum parameterization to anisotropic Multivariate 3D Gaussians with independent radii for each spatial direction (Fig~\ref{fig:overview}).

In Ultra-Wide-NeRF each sample at distance $t$ along a ray $(\mathbf{o}, \mathbf{d})$ is represented as a frustum segment $[t_j, t_{j+1}]_{j \in \{0,  \ \#samples\}}$ and encoded by a Gaussian $\mathcal{N}(\boldsymbol{\mu}, \boldsymbol{\mathrm{\Sigma}})$, with mean distance along the ray $\mu_t = \mu + \frac{2\mu \delta^2}{3\mu^2 + \delta^2}$ where $\mu = (t_j+t_{j+1})/2$ is the segment midpoint and $\delta = (t_{j+1}-t_j)/2$ is set to the segment's half-width.

\noindent\textbf{Variance decomposition.} The covariance $\boldsymbol{\mathrm{\Sigma}}$ is a full $3\times3$ matrix constructed from three distinct variances, encoding the spatial footprint of this region. Following~\cite{barron2021mip}, $\sigma^2_\text{ray}$ and $\sigma^2_\text{lateral}$ capture the along-ray and horizontally perpendicular-to-ray (lateral) variance. We additionally introduce $\sigma^2_\text{depth}$ to model the cross-plane elevational anisotropy of 3D US (vertically perpendicular-to-ray variance):
\begin{equation}
    \sigma^2_\text{ray} = \frac{\delta^2}{3} - \frac{4}{15} \cdot \frac{\delta^4(12\mu^2 - \delta^2)}{(3\mu^2 + \delta^2)^2} 
\end{equation} \label{eq:var_ray}
\begin{equation}
    \sigma^2_\text{\{lateral, depth\}} = r_\text{\{lat, dep\}}^2 \left(\frac{\mu^2}{4} + \frac{5\delta^2}{12} - \frac{4\delta^4}{15(3\mu^2 + \delta^2)}\right) \label{eq:var_lateraldepth} 
\end{equation}
where $r_\text{lat}$ and $r_\text{dep}$ are independent Gaussian base radii derived from the in-plane and elevational resolutions, $s_{lat}$ and $s_{dep}$, respectively, as $r_\text{\{lat, dep\}} = \frac{2 \cdot s_{\{lat, dep\}}}{\sqrt{12}}$. They represent the voxel's footprint in world coordinates, similar to pixels' footprint in \cite{barron2021mip}. (Fig. \ref{fig:overview})


\noindent\textbf{Covariance construction.}
The $3\times 3$ covariance matrix is assembled in a local orthonormal basis aligned with the ray geometry and then rotated to world coordinates:
\begin{equation}
    \boldsymbol{\mathrm{\Sigma}} = \mathbf{B}\, \text{diag}\!\left(\sigma^2_\text{ray},\; \sigma^2_\text{lateral},\; \sigma^2_\text{depth}\right) \mathbf{B}^\top
    \label{eq:covariance}
\end{equation}
where $\mathbf{B} = [\hat{\mathbf{d}},\; \mathbf{v}_1,\; \mathbf{v}_2] \in \mathbb{R}^{3\times 3}$ is an orthonormal basis with $\hat{\mathbf{d}}$ the ray direction, $\mathbf{v}_1$ a lateral vector (horizontally perpendicular to the ray, primarily in the imaging plane), and $\mathbf{v}_2$ a depth vector (perpendicular to both, aligned with the elevational direction).

\noindent\textbf{Integrated positional encoding.}
To reduce aliasing, we encode each frustum segment as a Gaussian footprint $\mathcal{N}(\boldsymbol{\mu}, \boldsymbol{\mathrm{\Sigma}})$ and apply Integrated Positional Encoding (IPE)~\cite{barron2021mip}.
Unlike standard positional encoding, which evaluates sine/cosine features at a point, IPE averages these features over the footprint.
This yields the usual multi-frequency features, but with frequency-dependent attenuation: high frequencies are suppressed when the footprint is large, while low frequencies are preserved.

In Ultra-Wide-NeRF we derive $\boldsymbol{\mathrm{\Sigma}}$ from the ray-, lateral-, and elevational (voluemetric depth) variances, and use the per-axis variances (the diagonal of $\boldsymbol{\mathrm{\Sigma}}$) to control this attenuation.
Because the footprint is anisotropic by construction, the attenuation is direction-dependent: frequencies are damped more along directions with lower resolution (larger variance) and preserved along directions with higher resolution.
This embeds the ultrasound sampling footprint directly into the network input representation and mitigates aliasing artifacts.


\subsection{3D-aware Loss Function}\label{ssec:loss}
All rays query a neural field in Cartesian world coordinates, resulting in voxel footprint in 3D space. We introduce 3D-informed loss that enforces 3D structural consistency via 3D loss functions, and Ultra-NeRF regularization \cite{wysocki2024ultra, Yesilkaynak_ultraNBA_2025}, applied to each full-stacked volume during training. The final loss is defined as:

\begin{equation}
    \mathcal{L} = (1 - \lambda)\, \mathcal{L}_\text{MSE}(\hat{V}, V) + \lambda\, \mathcal{L}_\text{SSIM} + \lambda_r \mathcal{L}_{\text{reg}}
    \label{eq:loss}
\end{equation}
where $\hat{V}$ is the rendered volume, $V$ is the ground truth volume, $\mathcal{L}_\text{MSE} = \|\hat{V} - V\|_2^2$ is the mean squared error, $\mathcal{L}_\text{SSIM} = 1 - \text{SSIM}(\hat{V}, V)$ is the structural similarity loss with a $3 \times 3$ Gaussian kernel, and $\lambda = 0.2$ controls the balance.
Additionally, after a warm-up period, we apply a finite-difference gradient
magnitude loss along all three spatial directions to preserve edge sharpness
and speckle texture:
\begin{equation}
    \mathcal{L}_\text{grad} = \bigl\| \|\nabla \hat{V}\|_2 - \|\nabla V\|_2 \bigr\|_1
\end{equation}
where $\nabla$ denotes the finite-difference gradient computed along the
$x$-, $y$-, and $z$-axes, and $\|\cdot\|_2$ is the per-voxel gradient magnitude.
\begin{figure}[t] 
    \centering
    \includegraphics[width=\textwidth]{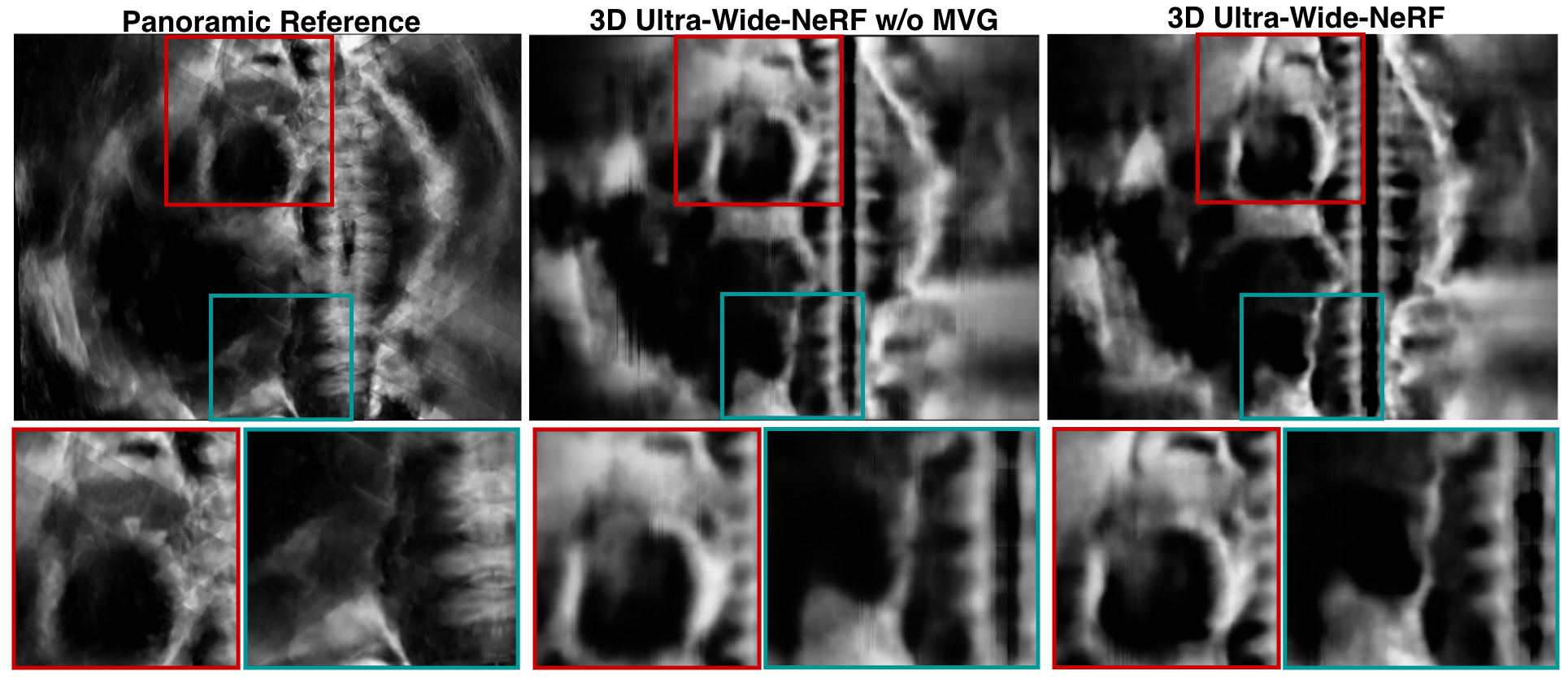} 
    \caption{\textbf{Comparison of WFOV Panoramic ICE Volume Reconstruction using Ultra-Wide-NeRF}. The first column shows the SOTA geometry-based reconstruction, the second one shows the voxel-based 3D Ultra-Wide-NeRF without MVG, and the last one shows the Ultra-Wide-NeRF using region estimates MVG. The second row illustrates close-up regions of stitching and aliasing artifacts removed by the region estimates.}
    \label{fig:results_comp}
\end{figure}
\section{Experiments and Results} 
\subsubsection{Data.}\label{ssec:exp_setup}
We evaluate Ultra-Wide-NeRF on two 3D ICE datasets, a static silicon phantom and an in vivo porcine acquisition.
Each dataset contains 3D overlapping ICE volume datasets. 
We compare Ultra-Wide-NeRF against registration based, geometry-aware panoramic reconstruction for ICE~\cite{hickler2024panoramic}. 
Both datasets are acquired with a \(360^{\circ}\) FoV ICE catheter (20 \(mm\) length, 90 \(mm\) radius) that enters the inferior vena cava (IVC), passes through the right atrium (RA), extends into the superior vena cava (SVC) and is then pulled back approximately 100-130 \(mm\). An acquisition produces 20 US volumes per second, creating a phantom dataset with 60 overlapping volumes and a porcine dataset with 17 volumes per cardiac phase, both with \(617\times617\times439\) voxels and isotropic spacing of \(0.3 mm\).
To assign a volume to a cardiac phase we use ECG gating.

\noindent\textbf{Network \& Training.} The network is an 8-layer MLP (256 hidden units, ReLU, 10 IPE frequency bands) minimising the loss in Sec.~\ref{ssec:loss}. We cast 250 conical frustra per volumetric depth over the $360^\circ$ convex definition, each integrated with 300 volumetric segments. The frustra start at the transducer and are estimated by MVGs whose principal axes encode the axial, lateral, and elevational resolutions of the ICE probe. The catheter FoV determines the lateral resolution, whereas the elevational resolution is scaled by the depth subsampling factor, directly embedding the transducer's anisotropic characteristics into sampling. We trained on an NVIDIA RTX 6000 Ada (49\,GB VRAM).
\subsection{Panoramic Reconstruction}
To validate the reconstruction fidelity and anti-aliasing capabilities of our approach, we benchmark Ultra-Wide-NeRF against the SOTA ICE panoramic compounding method~\cite{hickler2024panoramic}. In this evaluation, the SOTA baseline operates on full-resolution volumes (0.3 mm), whereas Ultra-Wide-NeRF is deliberately challenged to process volumes downsampled by a factor of 2 in the elevational direction (0.6 mm).
Visualized cross-sections (Fig.~\ref{fig:results_comp}) demonstrate that Ultra-Wide-NeRF recovers the identical anatomical context while producing sharper structural edges and eliminating the stitching artifacts inherent to the baseline. To isolate the contribution of our specific sampling strategy, we additionally compare Ultra-Wide-NeRF to an ablated model equipped with identical 3D spatial coherency and losses, but restricted to standard point sampling. This point-sampled ablation yields over-smoothed edges and washed-out structural details, explicitly validating that our continuous volumetric formulation is essential for preserving high-fidelity anatomical features, whilst being 10-20x faster to train for an accurate reconstruction, depending on the downsampling factor.
\begin{figure}[t] 
    \centering
    \includegraphics[width=\textwidth]{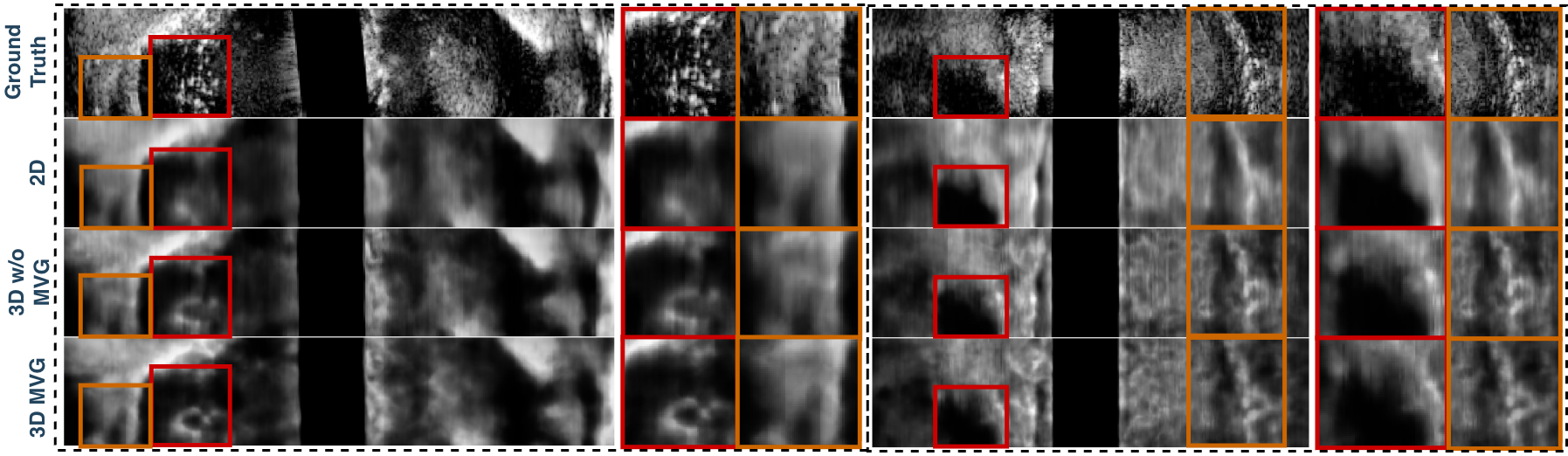} 
    \caption{\textbf{Qualitative comparison of held-out ICE volume reconstruction across anatomical planes.}
Each column shows a distinct anatomical plane with close-up regions highlighting aliasing artifacts.
Rows show, from top to bottom: ground truth, 2D Ultra-Wide-NeRF, 3D Ultra-Wide-NeRF w/o MVG, and \textbf{Ultra-Wide-NeRF with MVG}.}
    \label{fig:hold_out}
\end{figure}
\subsection{Novel View Synthesis}
We evaluate Ultra-Wide-NeRF's interpolation and novel-view synthesis capabilities across two scenarios:
Hold-out Volume Synthesis and Novel View Synthesis.
\noindent\textbf{Hold-out Volume Synthesis.} We evaluate Ultra-Wide-NeRF's ability to synthesize unobserved spatial context. We train the panoramic representation on a subset of volumes and test its synthesis capabilities on a strictly held-out volume bordered by two adjacent, overlapping training sweeps. As shown in Fig.~\ref{fig:hold_out}, the framework accurately reconstructs this unobserved volume directly from the learned panorama. The synthesized volume retains the anatomical details of the ground truth, confirming that our method delivers artifact-free novel view synthesis even when the acqusition steps along the physical acquisition trajectory are reduced.


 \begin{figure}[h] 
     \centering
     \includegraphics[width=0.9\textwidth]{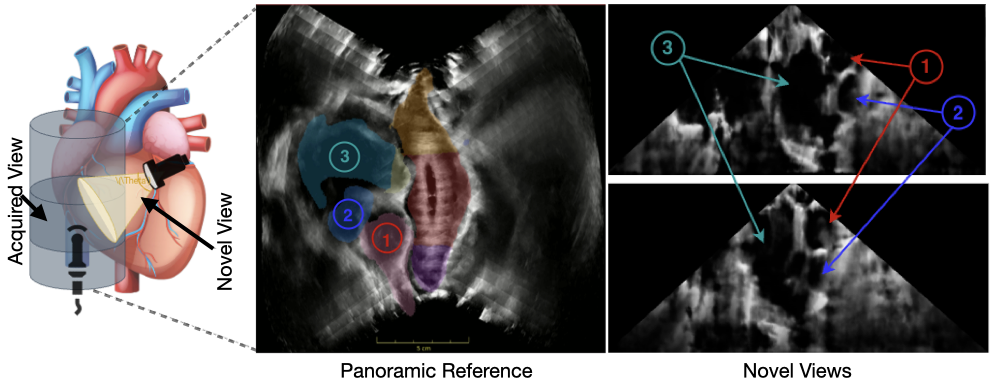} 
     \caption{\textbf{Novel view B-mode synthesis using Ultra-Wide-NeRF.} 
Centre: geometry-based panoramic reconstruction with overlaid segmentation masks. 
Right-top, bottom: synthesised novel views at $20^\circ$ convex angle, rendered from unseen probe poses outside the training trajectory.}
    \label{fig:novel_view_synthesis}
\end{figure}

\noindent\textbf{Novel Views Outside the Catheter Trajectory.} Beyond interpolating within the acquisition path, Ultra-Wide-NeRF enables synthesis ultrasound volumes from virtual viewpoints strictly outside the original transducer trajectory. Fig.~\ref{fig:novel_view_synthesis} shows that the model successfully synthesizes distinct off-trajectory B-mode images. By synthesising from entirely unobserved catheter positions Ultra-Wide-NeRF transforms the reconstructed volume into an interactive diagnostic tool. Clinicians can dynamically generate new perspectives, focus on specific anatomical regions of interest at higher resolutions, or utilize the virtual convex geometry to view the anatomy from optimal, unrecorded angles.



\section{Conclusion} 
We proposed Ultra-Wide-NeRF, bridging continuous neural representations with the diverging geometry of convex ultrasound. By explicitly modeling the spatial footprint of the acoustic beam using distance-dependent multivariate Gaussians, our framework acts as an inherent anti-aliasing filter to eliminate traditional stitching artifacts. Validations on phantom and porcine ICE datasets demonstrate superior artifact reduction and boundary enhancement compared to existing compounding methods. Delivering both artifact-free panoramic reconstructions and high-fidelity novel view synthesis, Ultra-Wide-NeRF expands critical spatial context from both acquired and virtual trajectories, establishing a robust foundation for seamless intraoperative navigation. While validated on intracardiac data, this geometric sampling strategy is inherently applicable to other clinical domains utilizing convex probes, including abdominal and obstetric imaging. By reliably expanding the observable field of view, this approach provides a practical tool to support more accurate spatial orientation during complex image-guided interventions.
\newpage
\bibliographystyle{splncs04}
\bibliography{Paper-0000}

@inproceedings{Yesilkaynak_ultraNBA_2025,
  title={UltraNBA Neural Bundle Adjustment for Pose Refinement in 3D Freehand Ultrasound},
  author={Yesilkaynak, V Bugra and Duque, Vanessa Gonzalez and Wysocki, Magdalena and Azampour, Mohammad Farid and Navab, Nassir and Mateus, Diana},
  booktitle={Proceedings of the IEEE/CVF International Conference on Computer Vision},
  pages={4363--4372},
  year={2025}
}

@article{poon2006three,
  title={Three dimensional ultrasound mosaicing},
  author={Poon, Timmy C and Rohling, Robert N},
  journal={IEEE Transactions on Medical Imaging},
  volume={25},
  number={5},
  pages={588--598},
  year={2006},
  publisher={IEEE}
}

@inproceedings{flach2016pure,
  title={PURE: panoramic ultrasound reconstruction by seamless stitching of volumes},
  author={Flach, Barbara and Makhinya, Maxim and Goksel, Orcun},
  booktitle={International Workshop on Simulation and Synthesis in Medical Imaging},
  pages={75--84},
  year={2016},
  organization={Springer}
}

@inproceedings{hickler2024panoramic,
  title={Panoramic anatomical context in 3D intracardiac echocardiography (ICE) with 3D registration and geometry based image fusion},
  author={Hickler, Julia and Neary Zajiczek, Lydia and Bandaru, Raja Sekhar and Balle S{\'a}nchez, Marcos and Hennersperger, Christoph and W{\"o}rz, Stefan},
  booktitle={Statistical Atlases and Computational Models of the Heart (STACOM)},
  pages={87--97},
  year={2024},
  organization={Springer}
}

@inproceedings{wysocki2024ultra,
  title={Ultra-NeRF: Neural Radiance Fields for Ultrasound Imaging},
  author={Wysocki, Magdalena and Azampour, Mohammad Farid and Eilers, Christine and Busam, Benjamin and Salehi, Mehrdad and Navab, Nassir},
  booktitle={Medical Imaging with Deep Learning (MIDL)},
  volume={227},
  pages={382--401},
  year={2024},
  organization={PMLR}
}

@inproceedings{dagli2024nerfus,
  title={NeRF US: Removing Ultrasound Imaging Artifacts from Neural Radiance Fields in the Wild},
  author={Dagli, Rishit and Hibi, Atsuhiro and Krishnan, Rahul G and Tyrrell, Pascal N},
  booktitle={Proceedings of Machine Learning Research},
  volume={252},
  pages={1--30},
  year={2024}
}

@article{fenster2001three,
  title={Three dimensional ultrasound imaging},
  author={Fenster, Aaron and Downey, Donal B and Cardinal, H Neale},
  journal={Physics in Medicine \& Biology},
  volume={46},
  number={5},
  pages={R67},
  year={2001},
  publisher={IOP Publishing}
}

@article{hsu2012real,
  title={Real time 4D ultrasound mosaicing and visualization},
  author={Hsu, Po Wei and Prager, Richard W and Gee, Andrew H and Treece, Graham M},
  journal={Medical Image Analysis},
  volume={16},
  number={3},
  pages={735--746},
  year={2012},
  publisher={Elsevier}
}

@book{cobbold2006foundations,
  title={Foundations of biomedical ultrasound},
  author={Cobbold, Richard S C},
  year={2006},
  publisher={Oxford University Press}
}

@book{prince2006medical,
  title={Medical imaging signals and systems},
  author={Prince, Jerry L and Links, Jonathan M},
  year={2006},
  publisher={Pearson Prentice Hall}
}

@inproceedings{barron2021mip,
  title={Mip-nerf: A multiscale representation for anti-aliasing neural radiance fields},
  author={Barron, Jonathan T and Mildenhall, Ben and Tancik, Matthew and Hedman, Peter and Martin-Brualla, Ricardo and Srinivasan, Pratul P},
  booktitle={Proceedings of the IEEE/CVF international conference on computer vision},
  pages={5855--5864},
  year={2021}
}

@article{guo2024ulre,
  title={Ulre-nerf: 3d ultrasound imaging through neural rendering with ultrasound reflection direction parameterization},
  author={Guo, Ziwen and Fang, Zi and Fu, Zhuang},
  journal={arXiv preprint arXiv:2408.00860},
  year={2024}
}

@inproceedings{salehi2015patient,
  title={Patient-specific 3D ultrasound simulation based on convolutional ray-tracing and appearance optimization},
  author={Salehi, Mehrdad and Ahmadi, Seyed-Ahmad and Prevost, Raphael and Navab, Nassir and Wein, Wolfgang},
  booktitle={International Conference on Medical Image Computing and Computer-Assisted Intervention},
  pages={510--518},
  year={2015},
  organization={Springer}
}

@article{burger2012real,
  title={Real-time GPU-based ultrasound simulation using deformable mesh models},
  author={Burger, Benny and Bettinghausen, Sascha and Radle, Matthias and Hesser, J{\"u}rgen},
  journal={IEEE transactions on medical imaging},
  volume={32},
  number={3},
  pages={609--618},
  year={2012},
  publisher={IEEE}
}

@inproceedings{duelmer2025ultraray,
  title={UltraRay: Introducing Full-Path Ray Tracing in Physics-Based Ultrasound Simulation},
  author={Duelmer, Felix and Azampour, Mohammad Farid and Wysocki, Magdalena and Navab, Nassir},
  booktitle={International Conference on Medical Image Computing and Computer-Assisted Intervention},
  pages={653--662},
  year={2025},
  organization={Springer}
}

\end{document}